# Virtual Rephotography: Novel View Prediction Error for 3D Reconstruction

MICHAEL WAECHTER[1], MATE BELJAN[1], SIMON FUHRMANN[1], NILS MOEHRLE[1], JOHANNES KOPF[2], and MICHAEL GOESELE[1]
[1]Technische Universität Darmstadt, [2]Facebook[*]

The ultimate goal of many image-based modeling systems is to render photo-realistic novel views of a scene without visible artifacts. Existing evaluation metrics and benchmarks focus mainly on the geometric accuracy of the reconstructed model, which is, however, a poor predictor of visual accuracy. Furthermore, using only geometric accuracy by itself does not allow evaluating systems that either lack a geometric scene representation or utilize coarse proxy geometry. Examples include light field or image-based rendering systems. We propose a unified evaluation approach based on novel view prediction error that is able to analyze the visual quality of *any* method that can render novel views from input images. One of the key advantages of this approach is that it does not require ground truth geometry. This dramatically simplifies the creation of test datasets and benchmarks. It also allows us to evaluate the quality of an unknown scene during the acquisition and reconstruction process, which is useful for acquisition planning. We evaluate our approach on a range of methods including standard geometry-plus-texture pipelines as well as image-based rendering techniques, compare it to existing geometry-based benchmarks, and demonstrate its utility for a range of use cases.

Categories and Subject Descriptors: I.3.7 [**Computer Graphics**]: Three-Dimensional Graphics and Realism

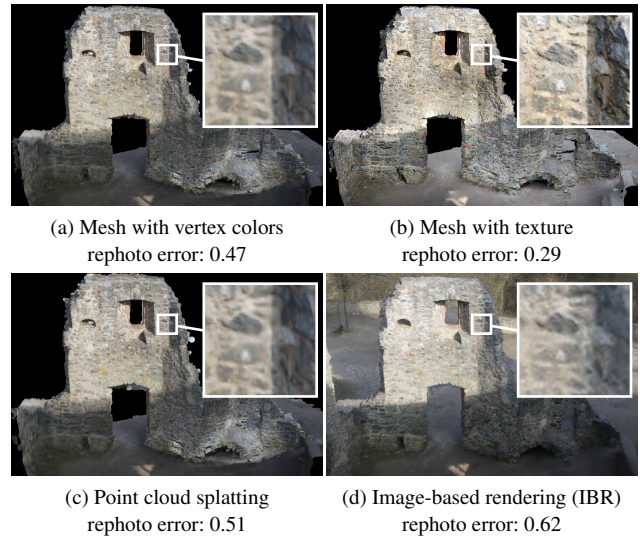

Fig. 1. Castle Ruin with different 3D reconstruction representations. Geometry-based evaluation methods [Seitz et al. 2006; Strecha et al. 2008; Jensen et al. 2014] cannot distinguish (a) from (b) as both have the same geometry. While the splatted point cloud (c) could in principle be evaluated with these methods, the IBR solution (d) cannot be evaluated at all.

(a) Mesh with vertex colors — rephoto error: 0.47
(b) Mesh with texture — rephoto error: 0.29
(c) Point cloud splatting — rephoto error: 0.51
(d) Image-based rendering (IBR) — rephoto error: 0.62

## 1. INTRODUCTION

Intense research in the computer vision and computer graphics communities has lead to a wealth of image-based modeling and rendering systems that take images as input, construct a model of the scene, and then create photo-realistic renderings for novel viewpoints. The computer vision community contributed tools such as structure from motion and multi-view stereo to acquire geometric models that can subsequently be textured. The computer graphics community proposed various geometry- or image-based rendering systems. Some of these, such as the Lumigraph [Gortler et al. 1996], synthesize novel views directly from the input images (plus a rough geometry approximation), producing photo-realistic results without relying on a detailed geometric model of the scene. Even though remarkable progress has been made in the area of modeling and rendering of real scenes, a wide range of issues remain, especially when dealing with complex datasets under uncontrolled conditions. In order to measure and track the progress of this ongoing research, it is essential to perform objective evaluations.

Existing evaluation efforts are focused on systems that acquire mesh models. They compare the reconstructed meshes with ground truth geometry and evaluate measures such as geometric completeness and accuracy [Seitz et al. 2006; Strecha et al. 2008; Jensen et al. 2014]. This approach falls short in several regards: First, only scenes with available ground-truth models can be analyzed. Ground-truth models are typically not available for large-scale reconstruction projects outside the lab such as PhotoCity [Tuite et al. 2011], but there is, nevertheless, a need to automatically evaluate reconstruction quality and to identify problematic scene parts. Second, evaluating representations other than meshes is problematic: point clouds can only be partially evaluated (only reconstruction accuracy can be measured but not completeness), and image-based rendering representations or light fields [Levoy and Hanrahan 1996] cannot be evaluated at all. And third, it fails to measure properties that are complementary to geometric accuracy. While there are a few applications where only geometric accuracy matters (e.g., reverse engineering or 3D printing), most applications that produce renderings for human consumption are arguably more concerned with visual quality. This is for instance the main focus in image-based rendering where the geometric proxy does not necessarily have to be accurate and only visual accuracy of the resulting renderings matters.

If we consider, e.g., multi-view stereo reconstruction pipelines, for which geometric evaluations such as the common multi-view benchmarks [Seitz et al. 2006; Strecha et al. 2008; Jensen et al. 2014] were designed, we can easily see that visual accuracy is complementary to geometric accuracy. The two measures are intrinsi-

---

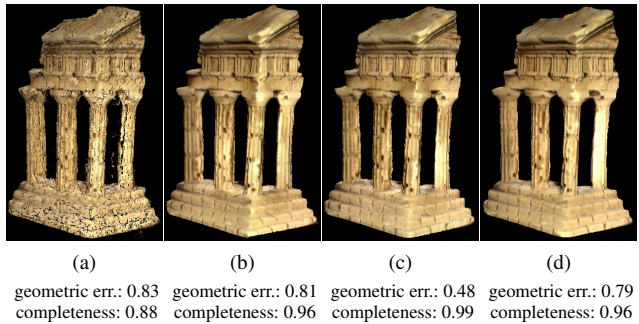

| (a) | (b) | (c) | (d) |
|---|---|---|---|
| geometric err.: 0.83 | geometric err.: 0.81 | geometric err.: 0.48 | geometric err.: 0.79 |
| completeness: 0.88 | completeness: 0.96 | completeness: 0.99 | completeness: 0.96 |

Fig. 2. Four Middlebury benchmark submissions: (a) Merrell [2007] Confidence and (b) Generalized-SSD [Calakli et al. 2012] have similar geometric error and different visual quality. (c) Campbell [2008] and (d) Hongxing [2010] have different geometric error and similar visual quality.

cally related since errors in the reconstructed geometry tend to be visible in renderings (e.g., if a wall's depth has not been correctly estimated, this may not be visible in a frontal view but definitely when looking at it at an angle), but they are not fully correlated and are therefore distinct measures. Evaluating the visual quality adds a new element to multi-view stereo reconstruction: recovering a good surface texture in addition to the scene geometry. Virtual scenes are only convincing if effort is put into texture acquisition, which is challenging, especially with datasets captured under uncontrolled, real-world conditions with varying exposure, illumination, or foreground clutter. If this is done well, the resulting texture may even hide small geometric errors.

A geometric reconstruction evaluation metric does not allow directly measuring the achieved visual quality of the textured model, and it is not always a good indirect predictor for visual quality: In Figure 2 we textured four submissions to the Middlebury benchmark using Waechter et al.'s approach [2014]. The textured model in Figure 2(a) is strongly fragmented while the model in 2(b) is not. Thus, the two renderings exhibit very different visual quality. Their similar geometric error, however, does not reflect this. Contrarily, 2(c) and 2(d) have very different geometric errors despite similar visual quality. Close inspection shows that 2(d) has a higher geometric error because its geometry is too smooth. This is hidden by the texture, at least from the viewpoints of the renderings. In both cases similarity of geometric error is a poor predictor for similarity of visual quality, clearly demonstrating that the purely distance-based Middlebury evaluation is by design unable to capture certain aspects of 3D reconstructions. Thus, a new methodology that evaluates visual reconstruction quality is needed.

3D reconstruction pipelines are usually modular: typically, structure from motion (SfM) is followed by dense reconstruction, texturing and rendering (in image-based rendering the latter two steps are combined). While each of these steps can be evaluated individually, our proposed approach is holistic and evaluates the complete pipeline *including* the rendering step by scoring the visual quality of the final renderings. This is more consistent with the way humans would assess quality: they are not concerned with the quality of intermediate representations (e.g., a 3D mesh model), but instead consider 2D projections, i.e., renderings of the final model from different viewpoints.

Leclerc et al. [2000] pointed out that in the real world inferences that people make from one viewpoint, are consistent with the observations from other viewpoints. Consequently, good models of real world scenes must be self-consistent as well. We exploit this self-consistency property as follows. We divide the set of captured images into a training set and an evaluation set. We then reconstruct the training set with an image-based modeling pipeline, render novel views with the camera parameters of the evaluation images, and compare those renderings with the evaluation images using selected image difference metrics.

This can be seen as a machine learning view on 3D reconstruction: Algorithms infer a model of the world based on training images and make predictions for evaluation images. For systems whose purpose is the production of realistic renderings our approach is the most natural evaluation scheme because it directly evaluates the output instead of internal models. In fact, our approach encourages that future reconstruction techniques take photo-realistic renderability into account. This line of thought is also advocated by Shan et al.'s Visual Turing Test [2013] or Vanhoey et al.'s appearance-preserving mesh simplification [2015].

Our work draws inspiration from Szeliski [1999], who proposed to use intermediate frames for optical flow evaluation. We extend this into a complete evaluation paradigm able to handle a diverse set of image-based modeling and rendering methods. Since the idea to imitate image poses resembles computational rephotography [Bae et al. 2010] as well as the older concept of repeat photography [Webb et al. 2010], we call our technique *virtual rephotography* and call renderings *rephotos*. By enabling the evaluation of visual accuracy, virtual rephotography puts visual accuracy on a level with geometric accuracy as a quality of 3D reconstructions.

In summary, our contributions are as follows:

—A flexible evaluation paradigm using the novel view prediction error that can be applied to any renderable scene representation,

—quantitative view-based error metrics in terms of image difference and completeness for the evaluation of photo-realistic renderings, and

—a thorough evaluation of our methodology on several datasets and with different reconstruction and rendering techniques.

Our approach has the following advantages over classical evaluation systems:

—It allows measuring aspects complementary to geometry, such as texture quality in a multi-view stereo plus texturing pipeline,

—it dramatically simplifies the creation of new benchmarking datasets since it does not require a ground truth geometry acquisition and vetting process,

—it enables direct error visualization and localization on the representation (see, e.g., Figure 11), which is useful for error analysis and acquisition guidance, and

—it makes reconstruction quality directly comparable across different scene representations (see, e.g., Figure 1) and thus closes a gap between computer vision and graphics techniques.

## 2. RELATED WORK

Photo-realistic reconstruction and rendering is a topic that spans both computer graphics and vision. Triggered (amongst other factors) by the Middlebury multi-view stereo benchmark [Seitz et al. 2006], work in the last decade has primarily focused on the reconstruction aspect, leading to many new reconstruction approaches. This paper takes a wider perspective by considering complete pipelines including the final rendering step.

In the following we first provide a general overview of evaluation techniques for 3D reconstructions, before we discuss image comparison metrics with respect to their suitability for our approach.

## 2.1 Evaluating 3D Reconstructions

Algorithms need to be objectively evaluated in order to *prove* that they advance their field [Förstner 1996]. For the special case of multi-view stereo (MVS) this is, e.g., done using the Middlebury MVS benchmark [Seitz et al. 2006]: It evaluates algorithms by comparing reconstructed geometry with scanned ground truth, and measures accuracy (distance of the mesh vertices to the ground truth) and completeness (percentage of mesh nodes within a threshold of the ground truth). The downsides of this purely geometric evaluation have been discussed in Section 1. In addition, this benchmark has aged and cannot capture most aspects of recent MVS research (e.g., preserving fine details when merging depth maps with drastically different scales or recovering texture). Strecha et al. [2008] released a more challenging benchmark with larger and more realistic architectural outdoor scenes and larger images. They use laser scanned ground truth geometry and compute measures similar to the Middlebury benchmark. Most recently, Jensen et al. [2014] released a more comprehensive dataset of controlled indoor scenes with larger, higher quality images, more accurate camera positions, much denser ground truth geometry and a modified evaluation protocol that is still based on geometric accuracy. They therefore address no fundamentally new challenges and all objections stated against Middlebury above apply here as well. Thus, researchers who address more challenging conditions or non-geometric aspects still have to rely mostly on *qualitative* comparison, letting readers judge their results by visual inspection.

Szeliski [1999] encountered the same problem in the evaluation of optical flow and stereo and proposed novel view prediction error as a solution: Instead of measuring how well algorithms estimate flow, he measures how well the estimated flow performs in frame rate doubling. Given two video frames for time $t$ and $t+2$, flow algorithms predict the frame $t+1$, and this is compared with the non-public ground truth frame. Among other metrics this has been implemented in the Middlebury flow benchmark [Baker et al. 2011]. Leclerc et al. [2000] use a related concept for stereo evaluation: They call a stereo algorithm self-consistent if its depth hypotheses for image $I_1$ are the same when inferred from image pairs $(I_0, I_1)$ and $(I_1, I_2)$. Szeliski's (and our) criterion is more flexible: It allows the internal model (a flow field for Szeliski, depth for Leclerc) to be wrong as long as the resulting rendering looks correct, a highly relevant case as demonstrated by Hofsetz et al. [2004]. Virtual rephotography is clearly related to these approaches. However, Szeliski only focused on view interpolation in stereo and optical flow. We extend novel view prediction error to the much more challenging general case of image-based modeling and rendering where views have to be extrapolated over much larger distances.

Novel view prediction has previously been used in image-based modeling and rendering, namely for BRDF recovery [Yu et al. 1999, Section 7.2.3]. However, they only showed qualitative comparisons. The same holds for the Visual Turing Test: Shan et al. [2013] ask users in a study to compare renderings and original images at different resolutions to obtain a *qualitative* judgment of realism. In contrast, we automate this process by comparing renderings and original images from the evaluation set using several image difference metrics to *quantitatively* measure and localize reconstruction defects. We suggest novel view prediction error as a general, quantitative evaluation method for the whole field of image-based modeling and rendering, and present a comprehensive evaluation to shed light on its usefulness for this purpose.

One method that similarly to ours does not require geometric ground truth data, is Hoppe et al.'s [2012] reconstruction quality metric for view planning in multi-view stereo. For a triangle mesh it checks each triangle's degree of visibility redundancy and maximal resolution. In contrast to our method it does not measure visual reconstruction quality itself, but it measures circumstances that are assumed to cause reconstruction errors.

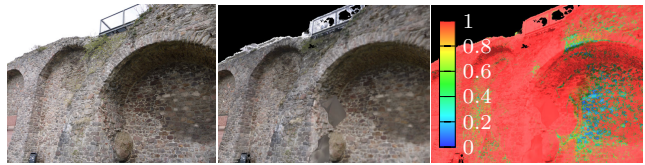

Fig. 3. *Left to right*: Input image, rendering from the same viewpoint, and HDR-VDP-2 result (color scale: VDP error detection probability)

## 2.2 Image Comparison Metrics

Our approach reduces the task of evaluating 3D reconstructions to comparing evaluation images with their rendered counterparts using image difference metrics. A basic image metric is the pixel-wise mean squared error (MSE) which is, however, sensitive to typical image transformations such as global luminance changes. Wang et al. [2004] proposed the structural similarity index (SSIM), which compares similarity in a local neighborhood around pixels after correcting for luminance and contrast differences. Stereo and optical flow methods also use metrics that compare image patches instead of individual pixels for increased stability: Normalized cross-correlation (NCC), Census [Zabih and Woodfill 1994] and zero-mean sum of squared differences are all to some degree invariant under low-frequency luminance and contrast changes.

From a conceptual point of view, image comparison in the photo-realistic rendering context should be performed with human perception in mind. The visual difference predictor VDP [Daly 1993] and HDR-VDP-2 [Mantiuk et al. 2011] are based on the human visual system and detect differences near the visibility threshold. Since reconstruction defects are typically significantly above this threshold, these metrics are too sensitive and unsuitable for our purpose. As shown in Figure 3, HDR-VDP-2 marks almost the whole rendering as defective compared to the input image. For high dynamic range imaging and tone mapping Aydın et al. [2008] introduced the dynamic range independent metric DRIM, which is invariant under exposure changes and highlights defective areas. However, DRIM's as well as HDR-VDP-2's range of values is hard to interpret in the context of reconstruction evaluation. Finally, the visual equivalence predictor [Ramanarayanan et al. 2007] measures whether two images have the same high-level appearance even in the presence of structural differences. However, knowledge about scene geometry and materials is required, which we explicitly want to avoid.

Given these limitations of perceptually-based methods, we utilize more basic pixel- or patch-based image correlation methods. Our experiments show that they work well in our context.

## 3. EVALUATION METHODOLOGY

Before introducing our own evaluation methodology, we would like to propose and discuss a set of general desiderata that should ideally be fulfilled: First, an evaluation methodology should *evaluate the actual use case* of the approach under examination. For example, if the purpose of a reconstruction is to record accurate geometric measurements, a classical geometry-based evaluation [Seitz et al. 2006; Strecha et al. 2008; Jensen et al. 2014] is the correct methodology. In contrast, if the reconstruction is used as a proxy in an image-based rendering setting, one should instead evaluate the achieved rendering quality. This directly implies that the results of different evaluation methods will typically not be consistent unless a reconstruction is perfect.

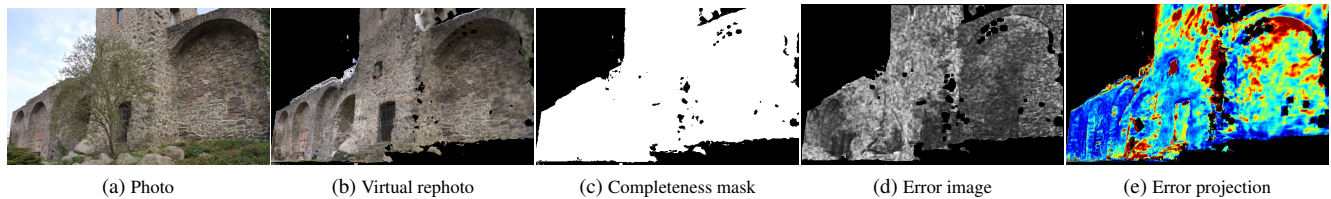

(a) Photo      (b) Virtual rephoto      (c) Completeness mask      (d) Error image      (e) Error projection

Fig. 4. An overview over the complete virtual rephotography pipeline.

Second, an evaluation methodology should be able to *operate solely on the data used for the reconstruction itself* without requiring additional data that is not available per se. It should be *applicable on site* while capturing new data. Furthermore, the *creation of new benchmark datasets should be efficiently possible*. One of they key problems of existing geometry-based evaluations is that the creation of ground-truth geometry models with high enough accuracy requires a lot of effort (e.g., high quality scanning techniques in a very controlled environment). Thus, it is costly and typically only used to benchmark an algorithm's behavior in a specific setting. An even more important limitation is that it cannot be applied to evaluate its performance on newly captured data.

Third, an evaluation methodology should *cover a large number of reconstruction methods*. The computer graphics community's achievements in image-based modeling and rendering are fruitful for the computer vision community and vice versa. Being able to only evaluate a limited set of reconstruction methods restricts intercomparability and exchange of ideas between the communities.

Finally, *the metric for the evaluation should be linear* (i.e., if the quality of model B is right in the middle of model A and C, its score should also be right in the middle). If this is not possible, the metric should at least give reconstructions an ordering (i.e., if model A is better than B, its error score should be lower than B's).

Fulfilling all of these desiderata simultaneously and completely is challenging. In fact, the classical geometry-based evaluation methods [Seitz et al. 2006; Strecha et al. 2008; Jensen et al. 2014] satisfy only the first and the last item above (in geometry-only scenarios it evaluates the use case and it is linear). In contrast, our virtual rephotography approach fulfills all except for the last requirement (it provides an ordering but is not necessarily linear). We therefore argue that it is an important contribution that is complementary to existing evaluation methodologies.

In the following, we first describe our method, the overall workflow and the used metrics in detail before evaluating our method in Section 4 using a set of controlled experiments.

### 3.1 Overview and Workflow

The key idea of our proposed evaluation methodology is that the performance of each algorithm stage is measured in terms of the impact on the final rendering result. This makes the specific system to be evaluated largely interchangeable, and allows evaluating different combinations of components end-to-end. The only requirement is that the system builds a model of the world from given input images and can then produce (photo-realistic) renderings for the view-points of test images.

We now give a brief overview over the complete workflow: Given a set of input images of a scene such as the one depicted in Figure 4a, we perform reconstruction using $n$-fold cross-validation. In each of the $n$ cross-validation instances we put $1/n$ th of the images into an evaluation set $E$ and the remaining images into a training set $T$. The training set $T$ is then handed to the reconstruction algorithm that produces a 3D representation of the scene. This could, e.g., be a textured mesh, a point cloud with vertex colors, or the internal representation of an image-based rendering approach such as the set of training images combined with a geometric proxy of the scene. In Sections 4 and 5 we show multiple examples of reconstruction algorithms which we used for evaluation.

The reconstruction algorithm then rephotographs the scene, i.e., renders it photo-realistically using its own, native rendering system with the exact extrinsic and intrinsic camera parameters of the images in $E$ (see Figure 4b for an example). If desired, this step can also provide a completeness mask that marks pixels not included in the rendered reconstruction (see Figure 4c). Note, that we regard obtaining the camera parameters as part of the reconstruction algorithm. However, since the test images are disjoint from the training images, camera calibration (e.g. using structure from motion [Snavely et al. 2006]) must be done beforehand to have all camera parameters in the same coordinate frame. State-of-the-art structure from motion systems are sub-pixel accurate for the most part (otherwise multi-view stereo would not work on them) and are assumed to be accurate enough for the purpose of this evaluation.

Next, we compare rephotos and test images with image difference metrics and ignore those image regions that the masks mark as unreconstructed (see Figure 4d for an example of a difference image). We also compute completeness as the fraction of valid mask pixels. We then average completeness and error scores over all rephotos to obtain global numerical scores for the whole dataset.

Finally, we can project the difference images onto the reconstructed model to visualize local reconstruction error (see Fig. 4e).

### 3.2 Accuracy and Completeness

In order to evaluate the visual accuracy of rephotos we measure their similarity to the test images using image difference metrics. The simplest choice would be the pixel-wise mean squared error. The obvious drawback is that it is not invariant to luminance changes. If we declared differences in luminance as a reconstruction error, we would effectively require all image-based reconstruction and rendering algorithms to bake illumination effects into their reconstructed models or produce them during rendering. However, currently only few reconstruction algorithms recover the true albedo and reflection properties of surfaces as well as the scene lighting (examples include Haber et al.'s [2009] and Shan et al.'s [2013] works). An evaluation metric that only works for such methods would have a very small scope. Furthermore, in real-world datasets illumination can vary among the input images and capturing the ground truth illumination for the test images would drastically complicate our approach. Thus, it seems adequate to use luminance-invariant image difference metrics. We therefore use the $YC_bC_r$ color space and sum up the absolute errors in the two chrominance channels. We call this $C_b+C_r$ error in the following.

This metric takes, however, only single pixels into consideration, mostly detects minor color noise, is unable to detect larger structural defects (demonstrated in Figure 10), and will be shown to *not* fulfill the ordering criterion defined in Section 3. Thus, we also analyze patch-based metrics, some of which are frequently used in stereo and optical flow: Zero-mean sum of squared differences (ZSSD), the structural dissimilarity index DSSIM = $(1-\text{SSIM})/2$ [Wang et al. 2004], normalized cross-correlation (we use 1-NCC instead of NCC to obtain a *dis*similarity metric), census [Zabih and Woodfill 1994], and the improved color image difference iCID [Preiss et al. 2014]. For presentation clarity throughout this paper we will only discuss $C_b+C_r$ error that does not fulfill the ordering criterion and 1-NCC error that does fulfill it. The other metrics exhibit very similar characteristics compared to 1-NCC and are shown in the full statistics in the supplemental material.

In conjunction with the above accuracy measures one must always compute some completeness measure, which states the fraction of the test set for which the algorithm made a prediction. Otherwise algorithms could resort to rendering only those scene parts where they are certain about their prediction's correctness. For the same reason machine learning authors report precision *and* recall and geometric reconstruction benchmarks [Seitz et al. 2006; Strecha et al. 2008] report reconstruction accuracy *and* completeness. For our purpose we use the percentage of rendered pixels as completeness. It is hereby understood that many techniques cannot reach a completeness score of 100 % since they do not model the complete scene visible in the input images.

## 4. EVALUATION

In the following, we perform an evaluation of our proposed methodology using a range of experiments. In Sections 4.1 and 4.2, we first demonstrate how degradations in the reconstructed model or the input data influence the computed accuracy. We show in particular, that our metric is not linear but fulfills the ordering criterion defined in Section 3. In Section 4.3 we discuss the relation between our methodology and the standard Middlebury MVS benchmark [Seitz et al. 2006]. Finally, we analyze in Section 4.4 to what extent deviating from the classical, strict separation of training and test sets decreases the validity of our evaluation methodology.

### 4.1 Evaluation with Synthetic Degradation

In this section we show that virtual rephotography fulfills the aforementioned ordering criterion on very controlled data. If we have a dataset's ground truth geometry and can provide a good quality texture, this should receive zero or a very small error. Introducing artificial defects into this model decreases the model's quality, which should in turn be detected by the virtual rephotography approach.

We take Strecha et al.'s [2008] Fountain-P11 dataset for which camera parameters and ground truth geometry are given. We compensate for exposure differences in the images (using exposure times from the images and an approximate response curve for the used camera) and project them onto the mesh to obtain a near-perfectly colored model with vertex colors (the resolution of the scanned ground truth mesh is so high that this is effectively equivalent to applying a high-resolution texture to the model). Rephotographing this colored model with precise camera calibration (including principal point offset and pixel aspect ratio) yields pixel-accurate rephotos.

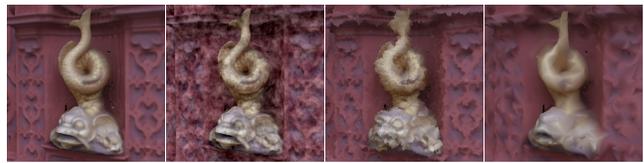

Fig. 5. Fountain-P11. *Left to right:* Ground truth, texture noise ($n_\text{tex} \approx 30\%$), geometry noise ($n_\text{geom} \approx 0.5\%$), and simplified mesh ($n_\text{simp} \approx 99\%$).

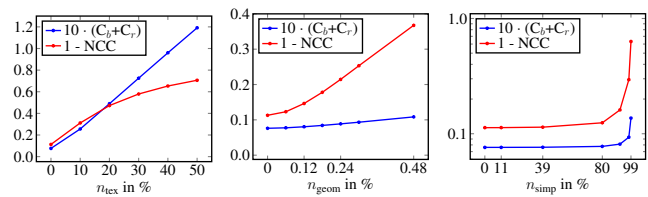

Fig. 6. 1-NCC and $C_b+C_r$ error for (*left to right*) texture noise, geometry noise and mesh simplification applied to Fountain-P11.

Starting from the colored ground truth we synthetically apply three different kinds of defects to the model (see Figure 5 for examples) to evaluate their effects on our metrics:

—Texture noise: In order to model random photometric distortions, we change vertex colors using simplex noise [Perlin 2002]. The noise parameter $n_\text{tex}$ is the maximum offset per RGB channel.
—Geometry noise: Geometric errors in reconstructions are hard to model. We therefore use a very general model and displace vertices along their normal using simplex noise. The parameter $n_\text{geom}$ is the maximum offset as a fraction of the scene's extent.
—Mesh simplification: To simulate different reconstruction resolutions, we simplify the mesh using edge collapse operations. The parameter $n_\text{simp}$ is the fraction of eliminated vertices.

In Fig. 6 we apply all three defects with increasing strength and evaluate the resulting meshes using our method. Both error metrics reflect the increase in noise with an increase in error—even the pixel-wise $C_b+C_r$ metric since all images were exposure-adjusted. One reason why the error does not vanish for $n_\text{tex}=n_\text{geom}=n_\text{simp}=0$, is that we cannot produce realistic shading effects easily.

### 4.2 Evaluation with Multi-View Stereo Data

In the following experiment we demonstrate the ordering criterion on more uncontrolled data, namely MVS reconstructions. Starting from the full set of training images at full resolution, we decrease reconstruction quality by (a) reducing the number of images used for reconstruction and (b) reducing the resolution of the images, and show that virtual rephotography detects these degradations.

We evaluate our system with two MVS reconstruction pipelines. In the first pipeline we generate a dense, colored point cloud with CMVS [Furukawa et al. 2010; Furukawa and Ponce 2010], mesh the points using Poisson Surface Reconstruction [Kazhdan et al. 2006], and remove superfluous triangles generated from low octree levels. We refer to this pipeline as CMVS+PSR. In the second pipeline we generate depth maps for all views with a general, community photo collection-based algorithm (CPC-MVS) [Goesele et al. 2007; Fuhrmann et al. 2015] and merge them into a global mesh using a multi-scale depth map fusion approach [Fuhrmann and Goesele 2011] to obtain a high-resolution output mesh with vertex colors. We refer to this pipeline as CPC+MS.

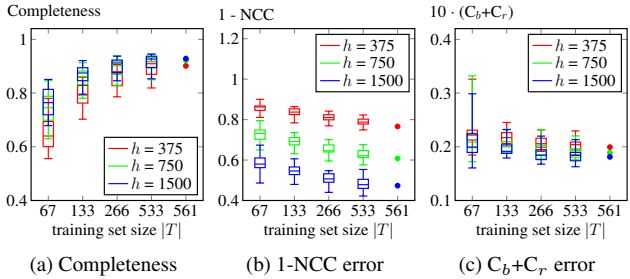

(a) Completeness  (b) 1-NCC error  (c) $C_b+C_r$ error

Fig. 7. City Wall virtual rephotography results for the CMVS+PSR pipeline. Boxplots show minimum, lower quartile, median, upper quartile, and maximum of 20 cross-validation iterations. Points for $|T|=561$ have been generated without cross-validation (see Section 4.4).

We use a challenging dataset of 561 photos of an old city wall with fine details, non-rigid parts (e.g., people and plants) and moderate illumination changes (it was captured over the course of two days)[1]. We apply SfM [Snavely et al. 2006] once to the complete dataset and use the recovered camera parameters for all subsets of training and test images. We then split all images into 533 training images ($T$) and 28 evaluation images ($E$). To incrementally reduce the number of images used for reconstruction we divide $T$ three times in half. We vary image resolution by using images of size $h \in \{375, 750, 1500\}$ ($h$ is the images' shorter side length). For evaluation we always render the reconstructed models with $h=750$ and use a patch size of $9 \times 9$ pixels for all patch-based metrics.

Figure 7 shows results for CMVS+PSR. CPC+MS behaves similarly and its results are shown in the supplemental material. The graphs give multiple insights: First (Figure 7a), giving the reconstruction algorithm more images, i.e., increasing $|T|$ increases our completeness measure. The image resolution $h$ on the other hand has only a small impact on completeness. Second (Figure 7b), increasing the image resolution $h$ decreases the 1-NCC error: If we look at the boxplots for a fixed $|T|$ and varying $h$, they are separated and ordered. Analogously, 1-NCC can distinguish between datasets of varying $|T|$ and fixed $h$: Datasets with identical $h$ and larger $|T|$ have a lower median error. These results clearly fulfill the desired ordering criterion stated above. Third (Figure 7c), the pixel-based $C_b+C_r$ error does not fulfill this criterion: Independent of $|T|$ and $h$ it assigns almost the same error to all reconstructions and is thus unsuitable for our purpose.

### 4.3 Comparison with Geometry-based Benchmark

The Middlebury MVS benchmark [Seitz et al. 2006] provides images of two objects, Temple and Dino, together with accurate camera parameters. We now investigate the correlation between virtual rephotography and Middlebury's geometry-based evaluation using the TempleRing and DinoRing variants of the datasets, which are fairly stable to reconstruct and are most frequently submitted for evaluation. With the permission of their respective submitters, we analyzed 41 TempleRing and 39 DinoRing submissions with publicly available geometric error scores. We transformed the models with the transformation matrices obtained from the Middlebury evaluation, removed superfluous geometry below the model base, textured the models [Waechter et al. 2014], and evaluated them.

Figure 8 shows 1-NCC rephoto error plotted against the geometric error scores. Analyzing the correlation between the two evalu-ations yields correlation coefficients between 1-NCC and geometric error of 0.63 for the TempleRing and 0.69 for the DinoRing, respectively. Figure 8 has, however, several outliers that deserve a closer look: E.g., the geometric errors for the methods Merrell Confidence [Merrell et al. 2007] (marked with an a) and Generalized-SSD [Calakli et al. 2012] (marked b) are similar whereas their 1-NCC errors differ strongly. Conversely, the geometric errors of Campbell [2008] (c) and Hongxing [2010] (d) are very different, but their 1-NCC error is identical. We showed renderings of these models in Figure 2 and discussed the visual dissimilarity of 2a and 2b and the similarity of 2c and 2d in Section 1. Apparently, visual accuracy seems to explain why for the pairs a/b and c/d the rephoto error does not follow the geometric error. Clearly virtual rephotography captures aspects complementary to the purely geometric Middlebury evaluation.

[1] http://www.gcc.tu-darmstadt.de/home/proj/mve/

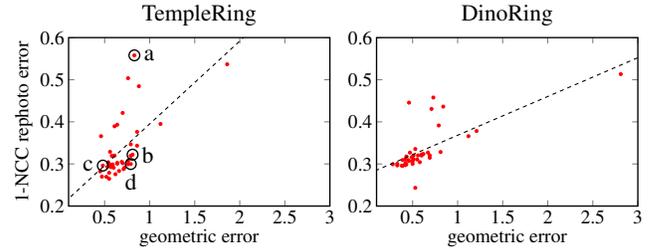

Fig. 8. 1-NCC rephoto error against geometric error for 41 TempleRing and 39 DinoRing datasets. Datasets marked with a–d are shown in Figure 2a–d. The dashed line is a linear regression fitted into the data points.

### 4.4 Disjointness of Training and Test Set

Internal models of reconstruction algorithms can be placed along a continuum between local and global methods. Global methods such as multi-view stereo plus surface reconstruction and texture mapping produce a single model that explains the underlying scene for all views as a whole. Local methods, such as image-based rendering, produce a set of local models (e.g., images plus corresponding depth maps as geometry proxies), each of which describes only a local aspect of the scene.

It is imperative for approaches with local models to separate training and test images since they could otherwise simply display the known test images for evaluation and receive a perfect score. We therefore randomly split the set of all images into disjoint *training* and *test sets* (as is generally done in machine learning) and use cross-validation to be robust to artifacts caused by unfortunate splitting. However, using all available images for reconstruction typically yields the best quality results and it may therefore be undesirable to "waste" perfectly good images by solely using them for the evaluation. This is particularly relevant for datasets which contain only few images to begin with and for which reconstruction may fail completely when removing images. Also, even though cross-validation is an established statistical tool, it is very resource- and time-consuming.

We now show for the global reconstruction methods CPC+MS and CMVS+PSR that evaluation can be done without cross-validation with no significant change in the results. On the City Wall dataset we omit cross-validation and obtain the data points for $|T|=561$ in Figure 7. Again, we only show the CMVS+PSR results here and show the CPC+MS results in the supplemental material. Most data points have a slightly smaller error than the median of the largest cross-validation experiments ($|T|=533$), which is reasonable as the algorithm has more images to reconstruct from.

Table I. Rephoto errors for different reconstruction representations of the Castle Ruin (Figure 1).

| Representation | Completeness | 1-NCC | $C_b+C_r$ |
| --- | --- | --- | --- |
| Mesh with texture | 0.66 | 0.29 | 0.017 |
| View-dependent texturing | 0.66 | 0.39 | 0.017 |
| Mesh with vertex color | 0.66 | 0.47 | 0.016 |
| Point cloud | 0.67 | 0.51 | 0.016 |
| Image-based rendering | 0.66 | 0.62 | 0.022 |

Neither CMVS+PSR nor CPC+MS seem to overfit the input images. We want to point out that, although this may not be generally applicable, it seems safe to not use cross-validation for the datasets and reconstruction approaches used here. In contrast, the image-based rendering approaches such as the Unstructured Lumigraph [Buehler et al. 2001] will just display the input images and thus break an evaluation without cross-validation.

## 5. APPLICATIONS

Here we show two important applications of virtual rephotography: In Section 5.1 we demonstrate our approach's versatility by applying it to different reconstruction representations. One possible use case for this is a future reconstruction quality benchmark that is open to all image-based reconstruction and rendering techniques. In Section 5.2 we use virtual rephotography to locally highlight defects in MVS reconstructions, which can, e.g., be used to guide users to regions where additional images need to be captured to improve the reconstruction.

### 5.1 Different Reconstruction Representations

One major advantage of our approach is that it handles arbitrary reconstruction representations as long as they can be rendered from novel views. We demonstrate this using the Castle Ruin dataset (286 images) and five different reconstruction representations, four of which are shown in Figure 1:

—Point cloud: Multi-view stereo algorithms like CMVS [Furukawa et al. 2010] output oriented point clouds, that can be rendered directly with surface splatting [Zwicker et al. 2001]. As splat radius we use the local point cloud density.
—Mesh with vertex color: This is typically the result of a surface reconstruction technique run on a point cloud.
—Mesh with texture: Meshes can be textured using the input images. Waechter et al. [2014] globally optimize the texture layout subject to view proximity, orthogonality and focus, and perform global luminance adjustment and local seam optimization.
—Image-based rendering: Using per-view geometry proxies (depth maps) we reproject all images into the novel view and render color images as well as per-pixel weights derived from a combination of angular error [Buehler et al. 2001] and TS3 error [Kopf et al. 2014]. We then fuse the color and weight image stack by computing the weighted per-channel median of the color images.
—View-dependent texturing: We turn the previous IBR algorithm into a view-dependent texturing algorithm by replacing the local geometric proxies with a globally reconstructed mesh.

Except for the IBR case we base all representations on the 3D mesh reconstructed with the CPC+MS pipeline. For IBR, we reconstruct per-view geometry (i.e., depth maps) for each input image using a standard MVS algorithm. For IBR as well as view-dependent texturing we perform leave-one-out cross-validation, the other representations are evaluated without cross-validation.

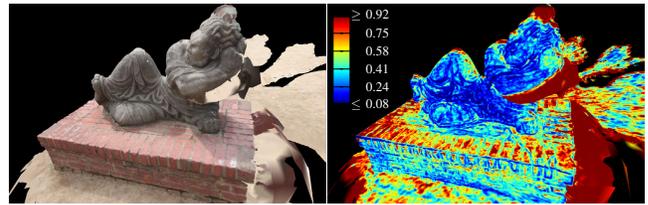

Fig. 9. Lying Statue: CMVS+PSR reconstruction and 1-NCC error projection (color scale: per vertex average 1-NCC error).

The resulting errors are shown in Table I. The pixel-based $C_b+C_r$ error is not very discriminative. The 1-NCC error on the other hand ranks the datasets in an order that is consistent with our visual intuition when manually examining the representations' rephotos: The textured mesh is ranked best. View-dependent texturing, that was ranked second, could in theory produce a lower error but our implementation does not perform luminance adjustments at texture seams (in contrast to the texturing algorithm), which are strongly detected as erroneous. The point cloud has almost the same error as the mesh with vertex colors since both are based on the same mesh and thus contain the same geometry and color information. Only their rendering algorithms differ, as the point cloud rendering algorithm discards vertex connectivity information. The image-based rendering algorithm is ranked lowest because our implementation suffers from strong artifacts caused by imperfectly reconstructed planar depth maps used as geometric proxies. We note, that our findings only hold for our implementations and parameter choices and no representation is in general superior to others.

### 5.2 Error Localization

If the evaluated reconstruction contains an explicit geometry model (which is, e.g., not the case for a traditional light field [Levoy and Hanrahan 1996]), we can project the computed error images onto the model to visualize reconstruction defects directly. Multiple error images projected to the same location are averaged. To improve visualization contrast we normalize all errors between the 2.5% and 97.5% percentile to the range [0, 1], clamp errors outside the range and map all values to colors using the "jet" color map.

Figure 9 shows a 1-NCC projection on a reconstruction of the Lying Statue dataset. It highlights blob-like Poisson Surface Reconstruction artifacts behind the bent arm, to the right of the pedestal, and above the head. Less pronounced, it highlights the ground and the pedestal's top which were photographed at acute angles.

Figure 10 shows that color defects resulting from combining images with different exposure are detected by patch-based metrics, but not by the pixel-based $C_b+C_r$ error as it cannot distinguish between per-pixel luminance changes due to noise and medium-scale changes due to illumination/exposure differences.

Figure 11 shows a textured City Wall model and its 1-NCC error projection. The error projection properly highlights hard to reconstruct geometry (grass on the ground or the tower's upper half, which was only photographed from a distance) and mistextured parts (branches on the tower or the pedestrian on the left). The supplemental material and the video show these results in more detail.

## 6. DISCUSSION

Evaluating the quality of a 3D reconstruction is a key component required in many areas. In this paper we propose an approach that focuses on the rendered quality of a reconstruction, or more precisely its ability to predict unseen views, without requiring a ground truth geometry model of the scene. It makes renderable reconstruction representations directly comparable and therefore accessible for a general reconstruction benchmark.

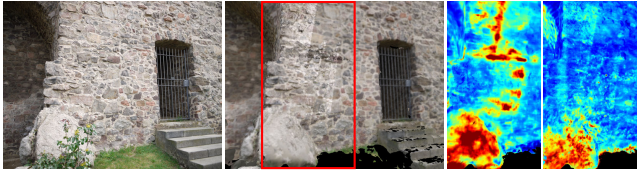

Fig. 10. *Left to right:* Photo, rephoto with color defect framed in red, 1-NCC and $C_b+C_r$ error projection of the framed segment.

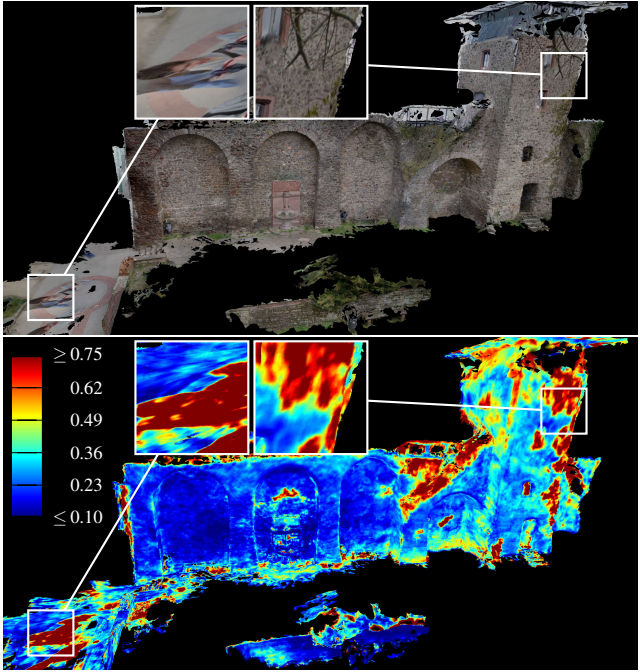

Fig. 11. Textured City Wall reconstruction. *Left:* A Pedestrian and a tree being used as texture. *Right:* The 1-NCC error projection highlights both.

While the high-level idea has already been successfully used for evaluation in other areas (optical flow [Szeliski 1999]) we are the first to use this for quantitative evaluation in image-based modeling and rendering and show with our experiments that it exhibits several desirable properties: First, it reports quality degradation when we introduce artificial defects into a model. Second, it reports degradation when we lower the number of images used for reconstruction or reduce their resolution. Third, it ranks different 3D reconstruction representations in an order that is consistent with our visual intuition. And fourth, its errors are correlated to the purely geometric Middlebury errors but also capture complementary aspects which are consistent with our visual intuition (Figure 2). Measuring visual accuracy solely from input images leads to useful results and enables many use cases, such as measuring the reconstruction quality improvement while tweaking parameters of a complete pipeline or while replacing pipeline parts with other algorithms. Highlighting reconstruction defects locally on the mesh can be used for acquisition planning.

Although we focused on the 1-NCC error throughout this paper, we note that all patch-based metrics (1-NCC, census, ZSSD, DSSIM, iCID) behave similarly: They all show a strictly monotonic error increase in the experiment from Section 4.1, and they are very close to scaled versions of 1-NCC in the experiments from Sections 4.2 and 4.3. In the supplemental material we show the full statistics for all experiments, all metrics, and the CPC+MS pipeline. The fact that all patch-based metrics behave largely similar does not come as a complete surprise: In the context of detecting global illumination and rendering artifacts Mantiuk [2013] "did not find evidence in [his] data set that any of the metrics [...] is significantly better than any other metric."

### 6.1 Limitations

An important limitation of our approach is the following: Since it has a holistic view on 3D reconstruction it cannot distinguish between different error sources. If a reconstruction is flawed, it can detect this but is unable to pinpoint what effect or which reconstruction step caused the problem. Thus—by construction—virtual rephotography does not replace but instead complements other evaluation metrics that focus on individual pipeline steps or reconstruction representations. For example, if we evaluate an MVS plus texturing pipeline which introduces an error in the MVS step, virtual rephotography can give an indication of the error since visual and geometric error are correlated (Section 4.3). Using error localization, i.e. by projecting the error images onto the geometric model, we can find and highlight geometric errors (Section 5.2). But to precisely debug only the MVS step, we have to resort to a geometric measure. Whether we use a specialized (Middlebury) or a general evaluation method (virtual rephotography) is a trade-off between the types and sources of errors the method can detect and its generality (and thus comparability of different reconstruction methods).

Another limitation revolves around various connected issues: Currently we do not evaluate whether a system handles surface albedo, BRDF, shading, interreflection, illumination or camera response curves correctly. They can in principle be evaluated, but one would have to use metrics that are less luminance-invariant than the investigated ones, e.g. mean squared error. Furthermore, one would need to measure and provide all information about the evaluation images which is necessary for correct novel-view prediction but cannot be inferred from the training images, e.g. illumination, exposure time, camera response curve, or even shadows cast by objects not visible in any of the images. Acquiring ground truth datasets for benchmarking, etc. would become significantly more complicated and time-consuming. In certain settings it may be appropriate to incorporate the above effects, but given that most 3D reconstruction systems do currently not consider them, and that virtual rephotography already enables a large number of new applications when using the metrics we investigated, we believe that our choice of metrics in this paper is appropriate for the time being.

### 6.2 Future Work

As already mentioned, our system's error projections could be used as input for next best view planning in autonomous exploration [Whaite and Ferrie 1997] or as live feedback to a user capturing a scene to guide him to acquire additional images of low quality areas. We would further like to conduct perceptual experiments to determine the extent to which our metrics correlate with human judgment about visual quality.

Furthermore we found, that datasets based on community photo collections (tourist photos from photo sharing sites) are very challenging for virtual rephotography because they exhibit great viewpoint or lighting variability, or occluders such as tourists. Most problematic here are test images with occluders or challenging il-lumination that are effectively impossible to predict by reconstruction algorithms. We achieve partial robustness towards this by (a) using image difference metrics that are (to a certain extent) luminance/contrast invariant, and (b) averaging all error images, that contribute to the error of a mesh vertex, during the error projection step. This can average out the contribution of challenging test images. In our experience, for some community photo collection datasets this seems to work and the errors highlighted by the 1-NCC projection partially correspond with errors we found during manual inspection of the reconstructions (e.g., in Figure 12 the 1-NCC projection detects dark, blurry texture from a nighttime image on the model's left side and heavily distorted texture on the model's right), while for others it does clearly not. In the future we would like to investigate means to make our evaluation scheme more robust with respect to this kind of datasets.

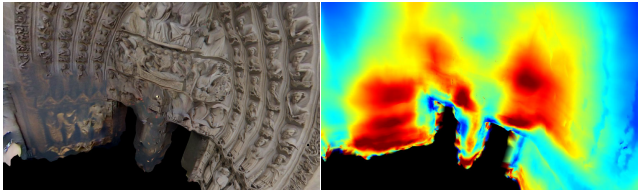

Fig. 12. Detail of a reconstruction (*left*) and its 1-NCC error projection (*right*) of Notre Dame based on 699 freely available Flickr images.

Finally, we plan to build a benchmark based on virtual rephotography that provides images with known camera parameters and asks submitters to render images using camera parameters of secret test images. For authors focusing on individual reconstruction pipeline aspects instead of a complete pipeline we plan to provide shortcuts such as an MVS reconstruction, depth maps for IBR, and texturing code. We believe that such a unified benchmark will be very fruitful since it provides a common base for image-based modeling and rendering methods from both the graphics and the vision community. It opens up novel and promising research directions by "allow[ing] us to measure progress in our field and motivat[ing] us to develop better algorithms" [Szeliski 1999].

ACKNOWLEDGMENTS

We are very grateful to Daniel Scharstein for providing us with the Middlebury benchmark submissions, the benchmark submitters who kindly permitted us to use their datasets, and Rick Szeliski for valuable discussions about this paper. This work was supported by the Intel Visual Computing Institute (project *RealityScan*).

...